\documentclass[journal,twoside,web]{ieeecolor2}

\usepackage{generic}
\usepackage{cite}
\usepackage{amsmath,amssymb,amsfonts}
\usepackage{algorithmic}
\usepackage{graphicx}
\usepackage{textcomp}
\usepackage[disable]{epstopdf}
\usepackage[inkscapelatex=false]{svg}

\begin{document}
\title{Effect of Performance Feedback Timing on Motor Learning for a Surgical Training Task}
\author{Mary Kate Gale, Kailana Baker-Matsuoka, Ilana Nisky, and Allison M. Okamura
\thanks{© 2025 IEEE. Personal use of this material is permitted. Permission from IEEE must be obtained for all other uses, in any current or future media, including reprinting/republishing this material for advertising or promotional purposes, creating new collective works, for resale or redistribution to servers or lists, or reuse of any copyrighted component of this work in other works.
}
\thanks{Received 08 July 2025. This work was supported in part by the National Science Foundation Graduate Research Fellowship, the Link Foundation Modeling, Simulation, and Training Fellowship, the National Science Foundation under Grant 1828993 and the U.S.-Israel Binational Science Foundation under Grant 2023022.}
\thanks{Mary Kate Gale, Kailana Baker-Matsuoka, and Allison Okamura are with the Departments of Bioengineering, Electrical Engineering, and Mechanical Engineering, respectively, at Stanford University, Stanford, CA 94305, USA (e-mail: mgale7@stanford.edu). }
\thanks{Ilana Nisky is with the Department of Biomedical Engineering at the Ben-Gurion University of the Negev, Beersheba, Israel.}
}

\maketitle

\newcommand{\responsetext}[1]{\textcolor{black}{#1}}
\bstctlcite{BSTcontrol}
\begin{abstract}
\textit{Objective:} Robot-assisted minimally invasive surgery (RMIS) has become the gold standard for a variety of surgical procedures, but the optimal method of training surgeons for RMIS is unknown. We hypothesized that real-time, rather than post-task, error feedback would better increase learning speed and reduce errors. 
\textit{Methods:} Forty-two surgical novices learned a virtual version of the ring-on-wire task, a canonical task in RMIS training. We investigated the impact of feedback timing with multi-sensory (haptic and visual) cues in three groups: (1) real-time error feedback, (2) trial replay with error feedback, and (3) no error feedback.
\textit{Results:} Participant performance was evaluated based on the accuracy of ring position and orientation during the task. Participants who received real-time feedback outperformed other groups in ring orientation. Additionally, participants who received feedback in replay outperformed participants who did not receive any error feedback on ring orientation during long, straight path sections. There were no significant differences between groups for ring position overall, but participants who received real-time feedback outperformed the other groups in positional accuracy on tightly curved path sections.
\textit{Conclusion:} The addition of real-time haptic and visual error feedback improves learning outcomes in a virtual surgical task over error feedback in replay or no error feedback at all.
\textit{Significance:} This work demonstrates that multi-sensory error feedback delivered in real time leads to better training outcomes as compared to the same feedback delivered after task completion. This novel method of training may enable surgical trainees to develop skills with greater speed and accuracy.

\end{abstract}

\vspace{1cm}
\section{Introduction}
\label{sec:introduction}
\IEEEPARstart{F}{or} many routine surgical procedures, robot-assisted minimally invasive surgery (RMIS) has become the state-of-the-art \cite{Maeso2010EfficacyMeta-analysis}\cite{Mjaess2023NewNow}. In RMIS, the surgeon remotely controls the movement of laparoscopic surgical tools inside the patient's body. The robot as an intermediary between the surgeon and patient allows for scaling and smoothing of surgeon gestures, as well as more ergonomic positioning for the surgeon during operation compared to standard open or laparoscopic surgery. Steerable stereoscopic cameras allow for three-dimensional visualization of patient tissue structures, improving surgeon vision and maneuverability over the flat images of traditional laparoscopic surgery \cite{Munz2004TheModels}. However, training surgeons for RMIS is not a perfected discipline, and the qualification process is not standardized nationally or globally \cite{Huffman2021AreProficiency}. This leads to research in characterization of both the parameters of expert RMIS use and how to accelerate novice users in their learning towards this mastery. Research in assessment of skill level has demonstrated that expert users tend to perform tasks more smoothly, quickly, and accurately with RMIS systems as compared to inexperienced or novice users \cite{Nisky2014EffectsSurgeons}\cite{Tausch2012ContentTracker}\cite{Sharon2021RateEvaluation}.

Prior work on motor learning and surgical training has aimed to improve learning outcomes through the addition of sensory augmentations, e.g., visual and haptic cues to guide users towards the most efficient or accurate method to execute a task. Visual augmentations, such as gradient color change based on task accuracy \cite{Enayati2018RoboticStudy}\cite{Oquendo2024HapticEnvironment}, have been demonstrated to improve RMIS task performance over time. Haptic augmentations are also employed in RMIS training research, both for physical guidance of performance through haptic force fields delivered by the surgeon-side manipulanda \cite{Oquendo2024HapticEnvironment}\cite{Coad2017TrainingSystem} and performance feedback through cutaneous skin-stretch \cite{Avraham2020TheAdaptation}\cite{Quek2019EvaluationTasks} and vibrotactile \cite{Sullivan2022HapticTask} feedback. 

Previous work on skill acquisition and retention in RMIS has focused on the manipulation of guidance and feedback for the acceleration of learning, \responsetext{and analysis and discussion have focused on the quantifiable learning outcomes rather than the underlying principles of motor learning driving the results of the work.} In the field of human motor neuroscience, motor learning of novel tasks can be divided into \textit{implicit learning}, which is the subconscious adaptation of a motion in response to predicted error of movement outcome, and \textit{explicit learning}, which involves the assessment of movement error after a motion is completed \cite{Krakauer2019MotorLearning}\cite{Masters2019AdvancesLearning}. Implicit motor learning is considered involuntary, and implicitly learned skills are recruited when subjects do not have time to react consciously to the parameters of a task \cite{Albert2021AnAdaptation}\cite{Mazzoni2006AnAdaptation}. Implicitly learned motor skills prove to be more stable over time and robust to task perturbation, stress, and multitasking \cite{Masters2008ImplicitMulti-tasking}\cite{El-Kishawi2021EffectEndodontics}. 

Motor adaptations to slight changes in a known task are thought to be primarily driven by implicit motor learning mechanisms \cite{Shadmehr1998Time-dependentSubjects}, while \textit{de novo} motor learning is initially dependent upon explicit motor learning processes \cite{Krakauer2019MotorLearning}\cite{Telgen2014MirrorNovo}. In the context of skill acquisition for RMIS, training paradigms that emphasize the implicit development of skills are desirable, due to the stable, robust nature of implicit motor learning. Additionally, RMIS trainees typically have some degree of surgical experience, which frames RMIS training as primarily a motor adaptation task as trainees transfer the known movements of surgery to the novel control system of the robot \cite{Nisky2014EffectsSurgeons}\cite{Nisky2014UncontrolledNovices} and learn to compensate for limited sensory input \cite{Okamura2009HapticSurgery}. Therefore, training for RMIS that emphasizes implicit motor learning processes may lead to superior skill acquisition and retention compared to training that emphasizes explicit motor learning processes. However, current curricula for RMIS training involve feedback about task performance after completion, which encourages conscious, or explicit, motor learning mechanisms \cite{Moorthy2003ObjectiveSurgery}\cite{Rahimi2024TrainingMethods}. \responsetext{Conversely, in most academic studies of surgical training, including \cite{Enayati2018RoboticStudy, Oquendo2024HapticEnvironment,Coad2017TrainingSystem,Avraham2020TheAdaptation,Quek2019EvaluationTasks,Sullivan2022HapticTask}, visual and haptic cues for feedback and guidance are delivered to participants in real time, which encourages subconscious, or implicit, motor learning. Some prior works, such as \cite{Enayati2018RoboticStudy}, provide additional pre- or post-task feedback, but this time-asynchronous feedback is intended to be complementary, not compared, to real-time guidance. Moreover, state-of-the-art studies in surgical training do not specifically investigate the types of learning that facilitate skill acquisition.} To allow us to compare the two styles of motor learning, we designed a novel set of haptic and visual augmentations for performance feedback in conjunction with a canonical RMIS training task, the ring-on-wire task \cite{Enayati2018RoboticStudy}\cite{Oquendo2024HapticEnvironment}\cite{Smith2014FundamentalsDevelopment}. Feedback delivered in real time was designed to train in the anticipation and correction of errors before they occur, emphasizing a style of implicit motor learning \cite{Mazzoni2006AnAdaptation}\cite{Schween2014OnlineRotation}, while feedback delivered after task completion was designed to train for post hoc error analysis and future correction, emphasizing explicit motor learning mechanisms \cite{Schween2017FeedbackRotation}. \responsetext{We provided information about ring position with haptic feedback and ring orientation with visual feedback, as described further in Section II.D. This combination of sensory augmentations was selected after pilot testing indicated it to be an intuitive and well-understood method of delivering error feedback. Our aim was to provide feedback that would be comprehensible and helpful irrespective of timing. Then, we examined \textit{how much more} effective that feedback became when delivery time was manipulated.} 

Additionally, we elected to present the experimental scene and deliver our multi-sensory feedback in a virtual reality (VR) environment, as opposed to traditional teleoperation of robotic arms within a physical environment. This was done to allow full characterization and control of all objects of interest within the experimental scene\responsetext{, as trackers for a physical scene are often subject to noise and error. The creation of a virtual scene also mimics common virtual training environments used in robotic surgery \cite{Rahimi2024TrainingMethods} and facilitates reproducibility.}

The contributions of this work are (1) a combination of vibrotactile haptic feedback and visual stimuli to deliver translational and rotational task cues simultaneously, and (2) the manipulation of feedback timing to promote differential learning outcomes based on targeting of implicit versus explicit motor learning mechanisms. We hypothesized that participants receiving real-time error feedback (targeting implicit learning) would outperform those receiving post hoc feedback (targeting explicit learning) or no error feedback. We also hypothesized that this difference would not be constant over space; i.e., there would be sections of the movement trajectories where more pronounced performance gaps would exist between groups. 

\begin{figure}[!t]
\centerline{\includegraphics[width=\columnwidth]{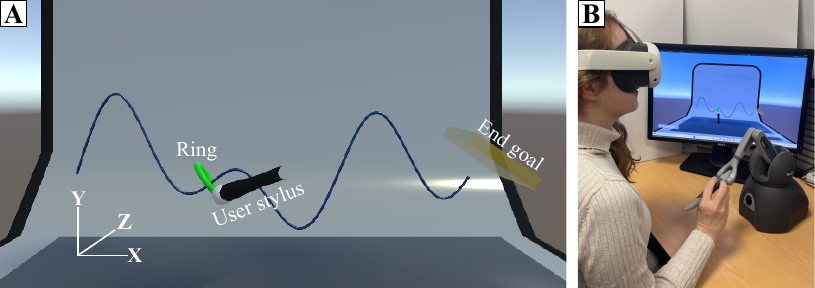}}
\caption{Experimental setup. (A) Virtual environment with salient objects labeled. (B) A user interacting with 3D Systems Touch device while the virtual environment is displayed on-screen for the experimenter.}
\label{experimental_setup}
\end{figure}

\section{Methods}
\subsection{Experimental Setup}
The experimental setup consisted of a desktop haptic device interfacing with a three-dimensional scene displayed in VR.
\subsubsection{Hardware and Software}
The virtual environment, as shown in Fig.~\ref{experimental_setup}A,  was designed and rendered in Unity (Unity Technologies, San Francisco, CA, USA). The scene was then displayed for subjects in VR on a Meta Quest 3 headset (Meta Platforms, Inc., Menlo Park, CA, USA) during experimentation. User movements were recorded and feedback was provided with a 3D Systems Touch haptic device (3D Systems Corporation, Rock Hill, SC, USA). The Touch device reads a 6-degree-of-freedom (DOF) pose from a user stylus and can supply 3-DOF of feedback force at up to 3.3 N. The device operates in a 0.431 $\times$ 0.348 $\times$ 0.165 m workspace. The Touch device interfaced with Unity using the Haptics Direct for Unity plugin (3D Systems Corporation), which allows for connection between the Touch device and Unity's built-in physics engine. This enables interactions between the Touch device and objects in a Unity scene.

\subsubsection{Virtual Scene and Data Collection}
The virtual scene consisted of a stage, a curved wire path in a two-dimensional plane, a ring, and an end zone (Fig.~\ref{experimental_setup}). The user moved the Touch stylus, which corresponded to on-screen stylus movement. Within the scene, two objects had physics interactions enabled: the stage floor and the ring. The stage floor was static and stiff, preventing both the ring and the user stylus from falling below the stage if dropped. The ring was dynamic and could be picked up by pressing a button on the Touch stylus when the virtual stylus tip came in contact with the ring. To simulate the typical lack of haptic feedback in an RMIS setting, once the ring was picked up, no haptic feedback was provided, such that subjects relied on visual cues only during manipulation. There were no modeled physics interactions between the ring and the wire path or the stylus and the wire path. Data were recorded at approximately 72 Hz and included the 6-DOF pose of the ring, the 6-DOF pose of the user stylus, the 3-DOF applied force from the Touch device, a Boolean of whether the ring was correctly on the path, and the time stamp. All data were processed and analyzed in MATLAB R2024b.

\subsection{Participants}
Fifty-three participants were recruited for the study (between 18 and 53 years of age; 25 female, 27 male, 1 non-binary; 51 right-handed). The protocol was approved by Stanford University's Internal Review Board (protocol \#22514), and all participants gave informed consent. Eleven participants were excluded from statistical analysis. Four participants were excluded due to technical errors with the system, one was excluded due to noncompliance with task instructions, three were excluded due to slow task performance (average trial length greater than 30 seconds), and three were excluded as performance outliers. \responsetext{Trial length was an exclusion criterion because the provided augmentations became irrelevant if participants completed the task as slowly as possible to avoid making mistakes, whether intentionally or unintentionally.} The threshold for exclusion of performance outliers was an improvement of the ring orientation score from baseline to post-test being greater than three standard deviations outside of the group mean. \responsetext{One participant from each group was removed as a performance outlier; two for being above group mean ($z = 3.61$; $z=3.61$) and and one for being below ($z=-3.61$.} After exclusions, there were 42 total participants, 14 in each of the three experimental groups. 

\subsection{Procedures}
During the experiment, participants were instructed to use the Touch device to move the ring along the wire path without contact between the two. They were asked to keep the orientation of the ring perpendicular to the wire path and to perform the task as quickly and accurately as possible. Participants were told that they would receive a performance score at the end of the experiment.

Participants donned the VR headset and first experienced a ``playground" with the stage, ring, and stylus in order to become comfortable with the Touch device controls. Once they were ready, participants moved on to the main experiment. There were 42 total trials\responsetext{: 5 pre-test, 30 main, 2 multitasking, and 5 post-test.}  One trial consisted of a complete movement along the path from the beginning position to the end zone. When the ring contacted the end zone, the trial ended and subjects saw a message instructing them to drop the ring and wait for the beginning of the next trial. 

The \textbf{pre-test trials} were unaugmented for all groups; that is, there was no information about performance or error other than what participants could see of the current state of the virtual scene and their understanding of task instructions. During the \textbf{main trials}, participants in the real-time and replay feedback groups experienced their respective augmentations, while participants in the control group continued to receive no information beyond their visual interpretation of the scene. Then, in the \textbf{multitasking trials}, augmentations were removed and participants were asked to complete a counting task (beginning with 100, count backwards by sevens out loud) in parallel with the ring-on-wire task. Multitasking trials were included to probe for the degree to which participants had internalized the motor skills necessary for the task; it was expected that participants who had learned the task more completely would perform more similarly in the multitasking trials to the ones immediately before and after, due to the relatively reduced mental load necessary for the ring-on-wire task \cite{El-Kishawi2021EffectEndodontics}\cite{Poolton2016MultitaskLearning}. \responsetext{We chose to include two multitasking trials to prevent additional confounding learning effects or fatigue while still providing some measurement reliability.} Finally, the \textbf{post-test trials} without augmentation assessed overall learning and final performance. Every ten trials, participants had a three-minute break of watching calming nature videos to prevent fatigue. \responsetext{The ten-trial interval was chosen as a way to evenly divide the experimental block with rest periods frequent enough to prevent participant fatigue. We played videos during the breaks to ensure that participants took advantage of the full break period; the videos were of natural scenes to be pleasant for the participants. These decisions were made through our refinement of the experimental setup during pilot testing.}
\begin{figure}[!t]
\centerline{\includegraphics[width=\columnwidth]{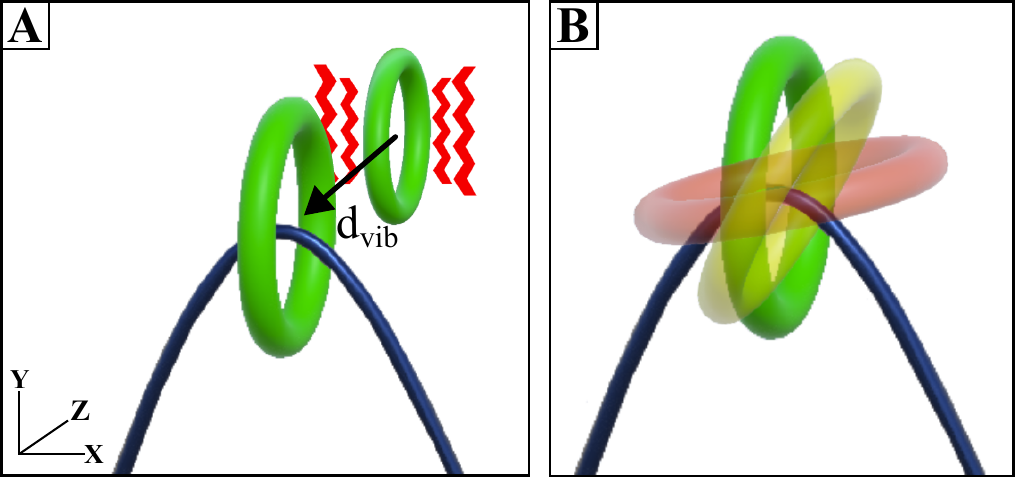}}
\caption{Sensory augmentations applied to real-time and replay feedback groups during the main trials. (A) Position error: when the ring contacted the wire or went off-path, participants would experience a vibration with constant frequency and amplitude until the ring returned to the wire path. \responsetext{$\text{d}_{\text{vib}}$ represents the direction of vibration in the Y-Z plane for the given position error.} (B) Rotation error: when the ring was correctly positioned, i.e., perpendicular to the wire path, it was green. When the ring rotated away from the correct angle, it moved through a gradient from yellow to red depending on magnitude of angle error only, with no directional information about error provided.}
\label{augmentations}
\end{figure}
\subsection{Augmentations}
During the main trials, participants in the augmentation groups (real-time and replay, respectively targeting implicit and explicit learning) received enhanced information about task performance via sensory augmentations (Fig.~\ref{augmentations}). Translational errors were cued by haptic augmentations and rotational errors were cued by visual augmentations.

The participants in the real-time group experienced these augmentations while they were completing the trial. The participants in the replay group experienced these augmentations immediately after the trial ended in the form of a trial replay including the augmentations. Participants in the replay group did not have sensory augmentations during the trial itself.
\subsubsection{Haptic Augmentation}
The haptic augmentation cued subjects to translational errors while completing the task (Fig.~\ref{augmentations}A). \responsetext{Because the ring was defined as an object in Unity, its known position and orientation allowed for computation of its translational accuracy along the wire path during the task.} When the inner edge of the ring contacted the wire or went off the path, subjects experienced a vibration of 200 Hz from the haptic device that continued until the ring was moved back onto the wire path. \responsetext{This vibration began essentially instantaneously, as allowed by the 72 Hz refresh rate of the physics engine.} The direction of the vibration was determined by the position error at the initial point that the ring left the wire path, that is,
\begin{equation}
    \text{d}_{\text{vibration}} =|0, y_{\text{i}} - y_{\text{e}}, z_{\text{i}} - z_{\text{e}}|,
\end{equation}
where $y_{\text{i}}$ and $z_{\text{i}}$ are the ideal position and $y_{\text{e}}$ and $z_{\text{e}}$ are the position of the center of the ring when the edge contacted the wire. The direction of vibration remained constant until error was corrected.

\subsubsection{Visual Augmentation}
The visual augmentation cued participants to rotational errors while completing the task (Fig.~\ref{augmentations}B). \responsetext{Similarly to the haptic augmentation, the position and orientation of the ring were computed within Unity; the desired orientation was calculated as a function of the ring's $x$-position and the curvature of the wire path at that point}. If the ring was oriented properly relative to its current location on the wire, i.e., exactly perpendicular, it was green. If the ring was oriented improperly, it changed colors through a continuous gradient from green through yellow and orange to red. The color of the ring at any moment was dependent on the magnitude of angle error from 0\textdegree \ to 90\textdegree, agnostic to the direction of error. During trials without sensory augmentations in all groups, the ring remained green during the duration of the task, irrespective of the accuracy of the ring rotation.

\subsection{Performance Metrics}
After data collection, user performance was assessed on trial time, translational path error, and rotational path error.
\subsubsection{Trial time} This was computed as the number of seconds from the first movement of the ring to the instant the ring contacted the end zone. This quantifies speed of task completion, which is an important metric in assessing surgical skill.
\subsubsection{Translational path error (TPE)} As in \cite{Oquendo2024HapticEnvironment}\cite{Coad2017TrainingSystem}, this metric was computed as the area of surface bounded by the current location of the center of the ring ($T_{n,C}$), the desired location of the center of the ring at the current time point ($T_{n,D}$), and the desired location of the center of the ring at the previous time point ($T_{n-1,D}$):
\begin{equation}
    \text{TPE} = \sum^N_{n=1} \text{Dist}[T_{n,D}, T_{n,C}]\ \text{Dist}[T_{n,D}, T_{n-1,D}]
\end{equation}
where $n$ is an individual sample point, $N$ is the total number of sample points in the trial, and $\text{Dist}[T_1, T_2]$ is the three-dimensional Euclidean distance between points $T_1$ and $T_2$. This metric represents ring ring position accuracy throughout a trial and its calculation is represented graphically in Fig.~\ref{metrics}A. \responsetext{TPE was selected as a performance metric due to its dependence on $\text{Dist}[T_{n,D}, T_{n-1,D}]$, the distance between the current point and the previously-sampled point along the path. Because trial duration and therefore number of samples varied both within and between subjects, longer trials would have arbitrarily higher error with an error metric dependent only upon $\sum \text{Dist}[T_{n,D}, T_{n,C}]$; by weighting each sample by $\text{Dist}[T_{n,D}, T_{n-1,D}]$, TPE is normalized and remains comparable across trials.}
\subsubsection{Rotational path error (RPE)} 
\begin{figure}[!t]
\centerline{\includegraphics[width=\columnwidth]{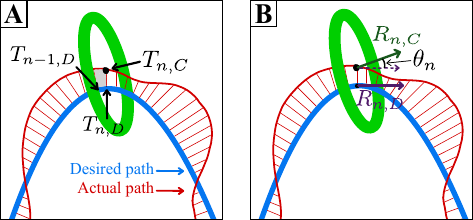}}
\caption{Computed quantities for translational path error (TPE) and rotational path error (RPE). (A) Current position ($T_{n,C}$), desired current position ($T_{n,D}$), and previous desired position ($T_{n-1,D}$). (B) Current pose ($R_{n,C}$), current desired pose ($R_{n,D}$), and angle between the two poses ($\theta_n$).}
\label{metrics}
\end{figure}

This metric was computed similarly to TPE, using the difference in angle between current normal vector to the ring ($R_{n,C}$) and desired normal vector to the ring ($R_{n,D}$):
\begin{equation}
\text{RPE} = \sum^N_{n=1} \theta_n\ \text{Dist}[T_{n,D}, T_{n-1,D}]
\end{equation}
where $\theta_n = \cos^{-1}\left(\frac{R_{n,C} \cdot R_{n,D}}{|R_{n,C}| |R_{n,D}|} \right)$, the angle between the current and desired normal vectors to the ring. This metric represents ring orientation accuracy throughout a trial and its calculation is represented graphically in Fig.~\ref{metrics}B. \responsetext{Similarly to TPE, RPE included the $\text{Dist}[T_{n,D}, T_{n-1,D}]$ term to normalize across trials of different lengths.}

\subsection{Statistical Analysis}
\responsetext{After data collection, 59 total trials were removed from the dataset: 54 trials where the ring was dropped and 5 where participants did not comply with instructions. This represented 3.3\% of all trials.} Additionally, the first 5\% by time of each trial was removed to account for start-up error associated with grasping and initially orienting the ring.

To determine the effects of different feedback paradigms on task performance, we used non-parametric statistical tests due to lack of underlying normality. The Kruskal-Wallis (KW) test was used to compare improvement in TPE, RPE, or trial time between baseline average and post-test average among the different groups for the overall path and specific sub-sections. When relevant, post-hoc comparisons using Dunn's test were used to compare differences between pairs of groups. To determine the effect of training on computed metrics within each group, the pairwise Wilcoxon signed-rank test was used. Significance was determined at a level of $p < 0.05$.

\begin{figure}[!t]
\centerline{\includegraphics[width=\columnwidth]{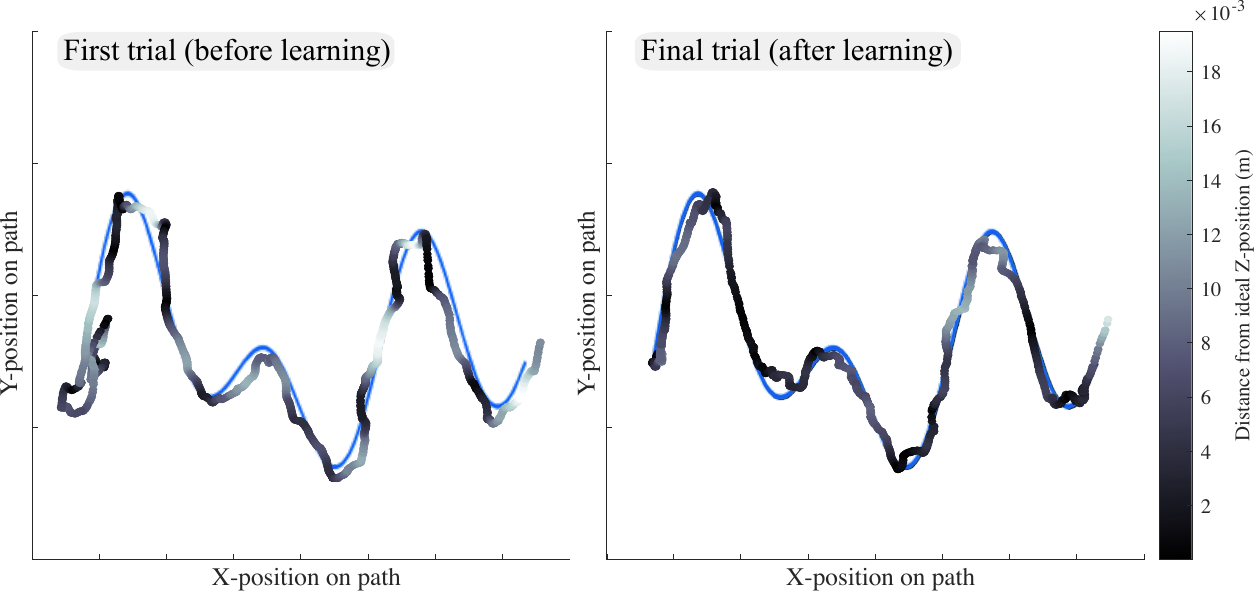}}
\caption{Translational path accuracy for first (left) and final (right) trials for sample participant in real-time feedback group. X- and Y-position of ring center are superimposed on true path (light blue). Absolute value of error in Z-position of ring center is indicated by path color, with light-colored sections having highest error. }
\label{path_training}
\end{figure}

\begin{figure}[!t]
\centerline{\includegraphics[width=\columnwidth]{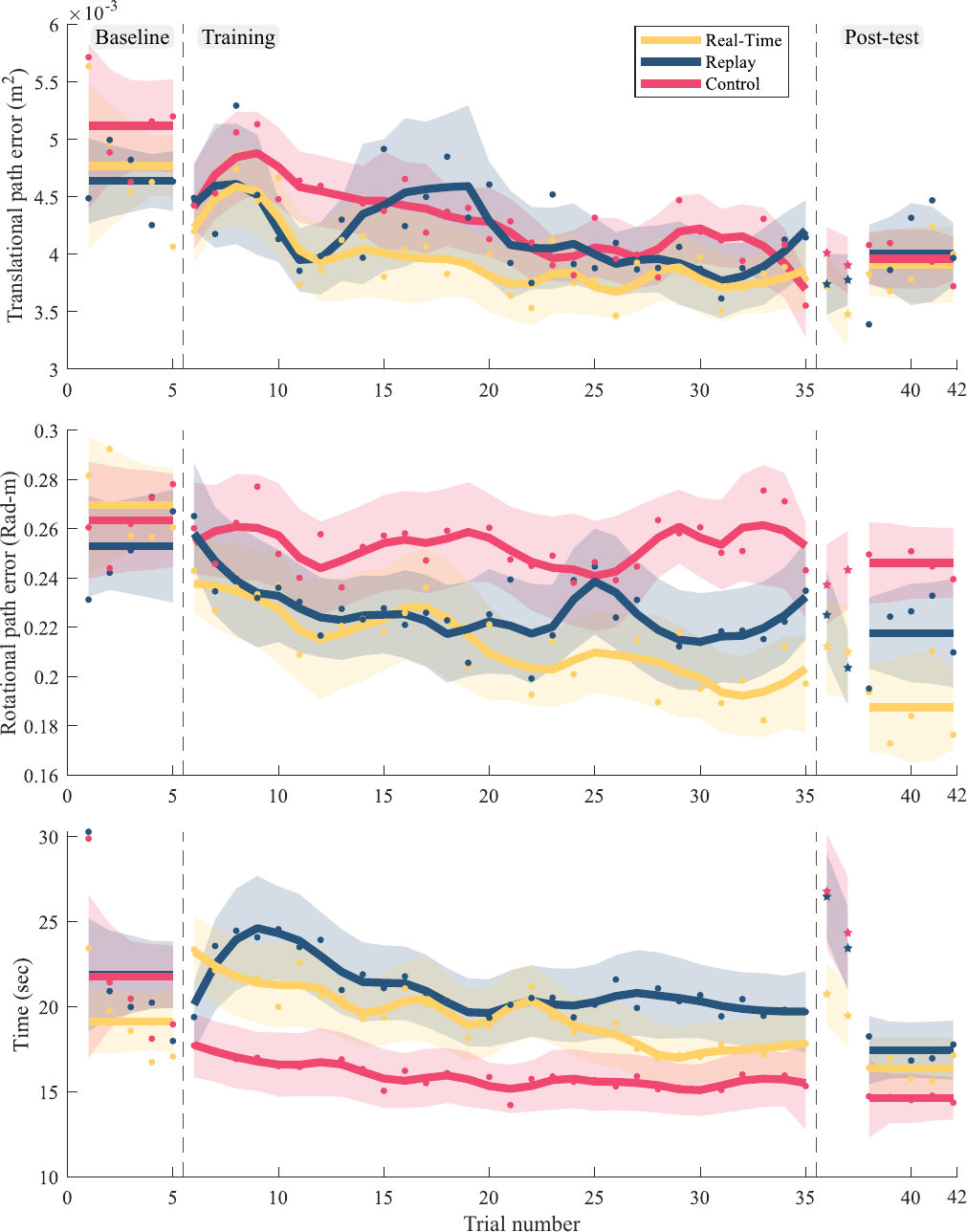}}
\caption{Learning curves for all three metrics over the course of the experiment for all three groups ($n =14$ per group, $n = 42$ total). Metrics were computed for each trial for each participant and then averaged across all participants in each group. Lines in first and last section represent baseline and post-test averages; line in middle represents locally estimated scatterplot smoothing curve. Trials marked by stars represent multitasking trials. Black vertical lines after trials 5 and 35 represent introduction and removal of augmentations. Shaded area represents $\pm1$ standard error of the mean.}
\label{learning_curves}
\end{figure}

\begin{figure*}[!t]
\centerline{\includegraphics[width=\textwidth]{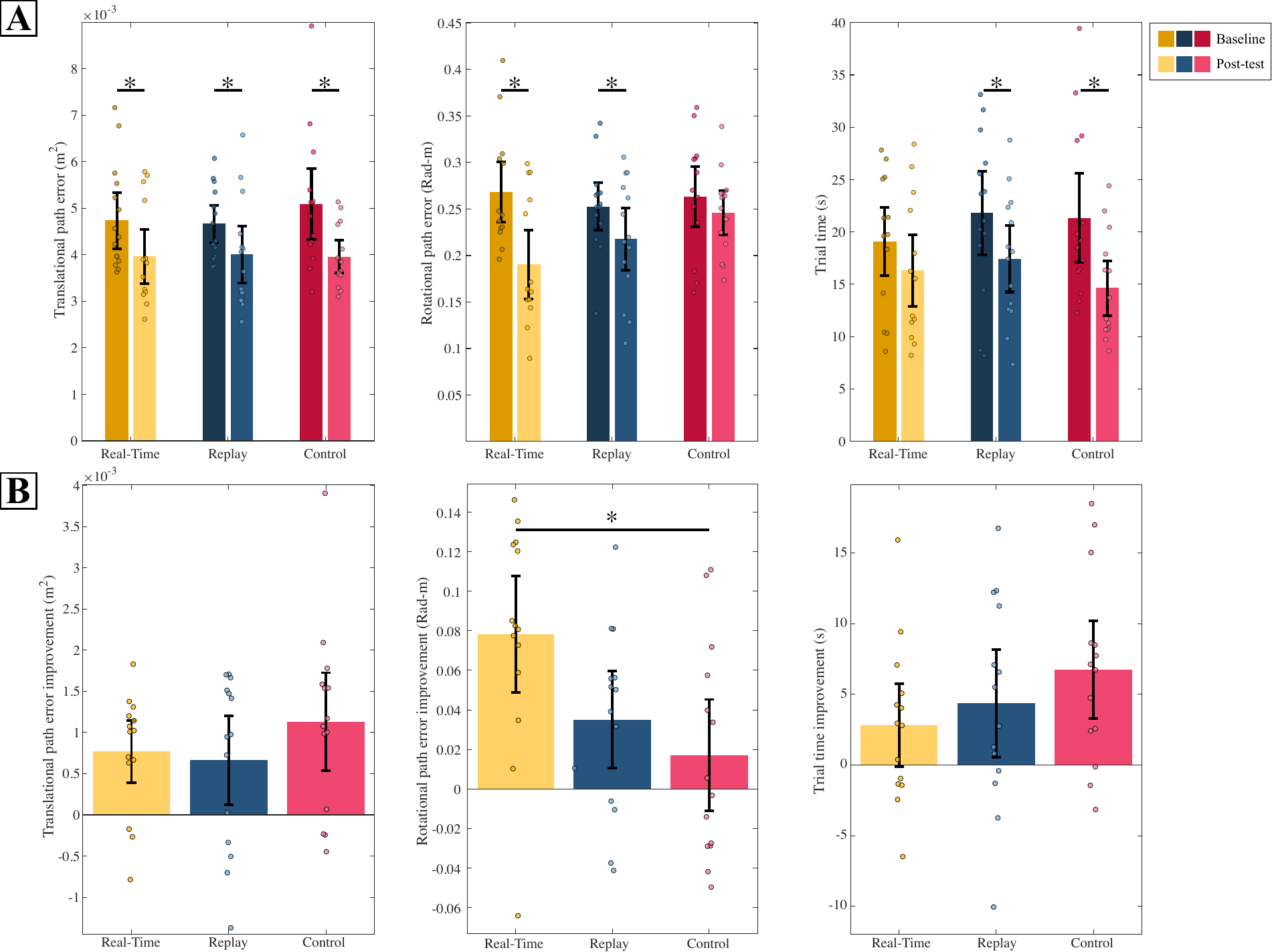}}
\caption{(A) Baseline (first 5 trials) and post-test (last 5 trials) averages for measured metrics within all groups ($n =14$ per group). Baseline bars are darker and post-test bars are lighter in color. Points represent average baseline and post-test performance for individual participants. Starred pairs represent statistically significant improvement in a metric within a certain group. Error bars represent 95\% confidence interval. (B) Improvement (baseline average minus post-test average) for each metric within each group, presented similarly to (A).}
\label{baseline_posttest}
\end{figure*}

\section{Results}

\subsection{Whole-Path Adaptation}
Over the course of training, participants learned to move the ring more smoothly and accurately along the wire path. Fig.~\ref{path_training} displays the translational accuracy for the first trial and last trial for a sample participant in the real-time feedback group. Before training, the movement of the ring fluctuated more in all three axes; after training, the participant was able to move the ring more steadily and with a trajectory that remained closer to the desired path. We formally examined the speed and shape of this learning process through the computation of learning curves. Fig.~\ref{learning_curves} shows these learning curves for three metrics (TPE, RPE, and completion time) calculated for each trial and averaged across participants. The learning curves display the average of the relevant metric at baseline, throughout the main trials, and during the post-test for each feedback group. On average, all groups improved performance on all metrics between baseline and post-test. The details of the statistical results are provided in Section~\ref{statistics}.

\subsubsection{Translational Path Error}
All participants improved slightly in their ability to keep the ring centered on the wire path throughout the experiment, but there was no significant effect of the presence or timing of feedback on this improvement. The learning curves for all participant groups displayed similar slopes and plateaued around the middle of the training trials, as seen in Fig.~\ref{learning_curves}. The translation component of the task was easier to comprehend and adhere to for most participants as compared to the rotation component of the task. Throughout the entire experiment, including baseline trials, all participants were on-path over 97\% of the time. 

\subsubsection{Rotational Path Error}
Participants in the real-time feedback group outperformed participants in other groups in rotational accuracy by the end of the experiment. Additionally, as seen in Fig.~\ref{learning_curves}, participants in the replay group did tend to outperform participants in the control group in rotational accuracy by the end of the experiment; however, due to high inter-subject variability in performance, this trend did not reach the required $p < 0.05$ level for significance. The participants in the replay group did significantly improve in rotational accuracy between the beginning and end of the experiment. Participants in the control group, who received no error feedback, remained largely constant in their ability to orient the ring properly and did not consistently improve in rotational accuracy throughout the course of the experiment.

\subsubsection{Trial Time}
Participants in the replay and control groups performed the task more quickly between the baseline and post-test measurements. Participants in the real-time feedback group did not significantly decrease their completion time, in part due to an already-low baseline completion time as compared to the other groups, but did tend towards faster task completion by the end of the experiment. Participants in groups that received feedback took longer per trial after feedback was first introduced in trial 6, but this was expected, as the new information gathered from the feedback made them more aware and cautious of their movements. By the end of the trials with feedback, participants in all groups had similar completion time per trial.

\subsection{Characterization of Performance Improvements} \label{statistics}
For all three primary metrics, overall performance improvement was computed as the difference between the average of that metric during the five post-test trials and the average of that metric during the five baseline trials.
\subsubsection{Within Groups}
Fig.~\ref{baseline_posttest}A shows the baseline performance (first 5 trials) and post-test performance (last 5 trials) for each group in the different metrics. Participants in all groups significantly reduced TPE per trial from the baseline to the post-test (real-time: $z=2.73;$ $p=0.006$, replay: $z=2.23;$ $p=0.026$, control: $z=2.73;$ $p=0.006$). Participants in groups receiving augmentation significantly improved in RPE per trial (real-time: $z=3.04;$ $p=0.002$, replay: $z=2.35;$ $p=0.019$), but participants in the control group did not ($z=0.91;$ $p=0.363$). Participants in the replay and control groups significantly decreased the time taken per trial between the baseline and post-test (replay: $z=1.98;$ $p=0.048$, control: $z=2.79;$ $p=0.005$) but participants in the real-time group did not ($z=1.79;$ $p=0.073$). 

\subsubsection{Between Groups}
To compare metric improvement between groups, we subtracted the average of the metric during the post-test from the average of the metric at baseline for each subject. Fig.~\ref{baseline_posttest}B presents the improvement between baseline and post-test performance in each metric for the experimental groups. There was no significant difference between groups in TPE improvement between baseline and post-test (KW $\chi ^2(2)=1.38;$ $p=0.501$). There was a difference between groups in RPE improvement (KW $\chi ^2(2)=8.47;$ $p=0.015$). Pairwise comparisons of RPE improvement demonstrated a significant difference between the real-time and control groups ($p=0.013$) but not between replay and real-time ($p=0.104$) or replay and control ($p=0.712$). There was no significant difference between groups in improvement in time per trial (KW $\chi ^2(2)=1.73;$ $p=0.421$).

\begin{figure*}[!t]
\centerline{\includegraphics[width=\textwidth]{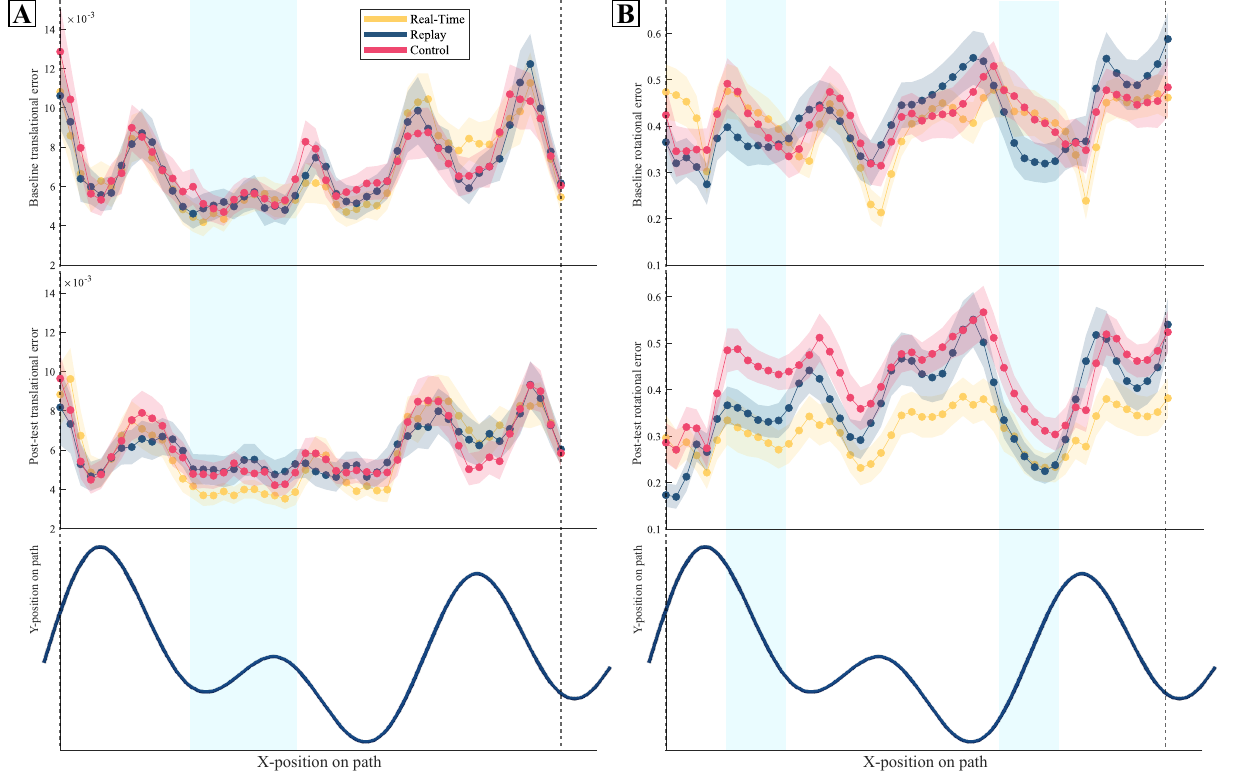}}
\caption{(A) Translational error at baseline (first 5 trials) and post-test (last 5 trials) over the course of the wire path (at bottom). Translational error was sampled at 50 evenly-spaced points along the path for each trial within each participant in a given trial range, then averaged across all participants in a group. Dashed black lines represent spatial locations of beginning and end of the graph on the wire path. Shaded areas represent locations of interest along the path. Fill represents 95\% confidence interval. (B) Rotational error at baseline (first 5 trials) and post-test (last 5 trials), computed and presented analogously to (A). }
\label{location_error}
\end{figure*}

\subsubsection{Multitasking Trials}
Multitasking trials were included after the primary training block but before the post-test trials. There were no statistically significant differences between groups in TPE (KW $\chi ^2(2) = 2.59;$ $p=0.274$), RPE (KW $\chi ^2(2)=2.94;$ $p=0.230$), or elapsed time (KW $\chi ^2(2)=1.41;$ $p=0.495$) during multitasking trials, and no differences between TPE and RPE in multitasking trials and the immediately preceding and following trials. In all groups, trial time was higher during the multitasking trials than during the surrounding trials.

\subsection{Performance Dependence on Task Geometry}
To further examine participant behavior during the course of a trial, we computed error at specific locations along the path of the wire. Translational error was the Euclidean distance between the desired and current position (i.e., $\text{Dist}[T_{n,C},T_{n,D}]$) and rotational error was the angle difference between the current and desired pose (i.e., the $\theta_n$ term from RPE). Because we were interested in instantaneous, location-specific error and not participant speed, we excluded the dependency on the previous time point (as found in TPE and RPE) when calculating this error. Fig.~\ref{location_error}A demonstrates translational error sampled at 50 evenly-spaced points along the path for the different groups at baseline and post-test, while Fig.~\ref{location_error}B demonstrates rotational error sampled at 50 points along the path for different groups at baseline and post-test.

Fig.~\ref{location_error} shows differences in performance between participant groups at particular points along the wire path. Specifically, participant groups were differentiable in translational error in the tighter curves in the second quarter of the path, and in rotational error in the long, straight sections.

\begin{figure*}[!t]
\centerline{\includegraphics[width=\textwidth]{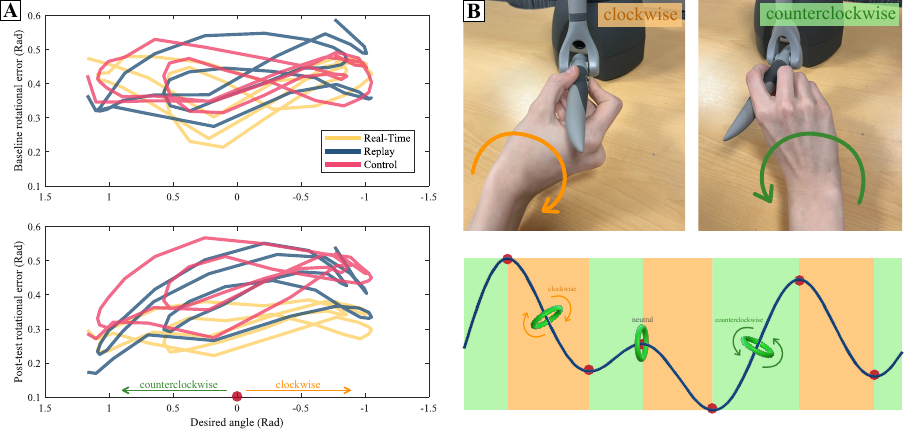}}
\caption{(A) Average rotational error for all participants in a group for first 5 trials (top) and last 5 trials (bottom) compared to desired angle. Rotational error was calculated at 50 evenly-spaced points throughout the path for each trial in the range for each participant. Points were connected to represent the rotational error vs. desired angle throughout the entire motion of a trial.  (B) Demonstration of clockwise (orange) and counterclockwise (green) wrist motions and corresponding sections on the wire path.}
\label{angle_vs_error}
\end{figure*}

\subsubsection{Tight Curves}
On the whole path, there were negligible differences between participant groups. However, as shown in Fig.~\ref{location_error}A, on the portion of the path with the most rapid changes in direction, participants in the real-time feedback group slightly outperformed those in the replay and control groups in translational error. This indicates that real-time feedback assisted in attentiveness to correct positioning in areas where rapid directional changes were required, but such information was not necessary or helpful in stretches with relatively long distances between directional changes.

Additionally, in all groups, translational error tended to be higher on long, straight sections and lower at inflection points. This may be due to participants moving the ring more quickly through long, straight areas that did not require significant rotational changes, leading to higher translational error, and being more careful around inflection points due to necessary rotational changes, leading to lower translational error.
\subsubsection{Long Straightaways}
In general, participants in the real-time feedback group outperformed other participants in rotational accuracy throughout the majority of the path. However, as seen in Fig.~\ref{location_error}B, participants in both groups receiving feedback (real-time and replay) outperformed the control group in rotational error on long, straight sections of the path. This indicates that the introduction of post hoc feedback in the replay group was more assistive in rotational performance during path sections where the same ring angle was maintained for a longer period of time, as compared to the rapid changes needed in the tighter curves elsewhere on the path.

\subsection{Error vs. Desired Angle}
Participants in the real-time group had the lowest RPE during post-test as compared to participants in the replay group and control group, and participants in both feedback groups outperformed the control group on straight sections of the path. These long, straight sections, such as those highlighted in Fig.~\ref{location_error}B, correspond to periods of relatively extreme and prolonged wrist rotation. This difference was most pronounced at angles requiring a clockwise wrist rotation. Fig.~\ref{angle_vs_error} demonstrates rotational error as a function of desired angle at baseline and in post-test for the three participant groups. In Fig.~\ref{angle_vs_error}A, at baseline, participants in all groups had relatively consistent error across all desired angles. By the post-test, participants in the control group and, to a lesser extent, the replay group, tended to favor counterclockwise angles (which correspond to upward slopes as the ring is moved left to right), while participants in the real-time group had high accuracy at both counterclockwise and clockwise angles. 

Typically, participants grasped the ring in an initially neutral position with anticipation of the first slope, which requires a counterclockwise rotation. Because of this initial grasp, which often did not anticipate later necessary clockwise rotations, participants had higher angle error on sections of the path requiring clockwise rotations. Participants who received error feedback in real-time were made aware of this tendency towards higher error in clockwise rotation and were able to overcome the natural inclination towards insufficient rotation.

\section{Discussion}
We demonstrated that training inspired by the principles of implicit motor learning (real-time error feedback) leads to superior performance outcomes as compared to training inspired by explicit motor learning (replay error feedback). 

Participants were able to learn the translational component unaided by error feedback. This might be due to the relatively large size of the ring compared to the wire path and the ease of depth perception afforded by the virtual reality setting. Participants who received real-time feedback outperformed those who did not in translation accuracy on the most technically difficult section of the path. This indicates that for tasks with higher overall difficulty, participants receiving real-time feedback may outperform those receiving replay or no error feedback in translational accuracy throughout the task.

Over the entire path, participants in the real-time feedback group outperformed other participants in rotational accuracy, as measured by RPE, by the end of training. Participants in the replay feedback group also improved in RPE by the end of the training period; conversely, participants receiving no error feedback did not significantly improve in RPE between baseline and post-test. Additionally, participants in both feedback groups outperformed participants who did not receive feedback on long, straight path sections where a particular rotation was maintained for a longer period of time, and participants in the real-time feedback group outperformed all other participants in path sections that required extreme clockwise rotation.

All participants reduced the amount of time necessary to complete a trial between baseline and post-test. In the two groups receiving error feedback, the amount of time required per trial increased temporarily during the first five trials of the training process due to the added complication of integrating external feedback into task completion; by the post-test trials at the end, these differences had subsided and all groups had similar task completion times. 

\subsection{Dissociation of Implicit and Explicit Motor Learning}
We demonstrated that participants who learned a task under a training paradigm inspired by the principles of implicit motor learning outperformed those who learned under a paradigm inspired by explicit motor learning. However, we did not separate the contribution of the two learning processes; i.e., the amount of performance improvement due to subconscious implicit processes versus the amount due to conscious explicit effort. Because implicit and explicit motor learning are closely intertwined and often both active in novel skill acquisition and known skill adaptation, it can be difficult to disentangle the learning contributions of the two mechanisms \cite{Kleynen2014UsingLearning}.

In the inclusion of multitasking trials after the completion of training, we anticipated that participants in the real-time group, whose feedback paradigm was inspired by implicit learning strategies, would retain performance at a higher rate in comparison to the participants in the replay and control groups. This expectation was based on the addition of conscious effort that would be required for a new cognitively demanding task, which would cause a breakdown in strategic components for an explicitly-learned skill \cite{Krakauer2019MotorLearning}\cite{McDougle2015ExplicitLearning}. 

However, there were no differences in accuracy-based metrics (i.e., TPE and RPE) between the multitasking trials and the trials immediately preceding and following within groups, and no differences in these accuracy-based metrics during multitasking trials between groups. Conversely, the time taken for the multitasking trials was significantly higher than the surrounding trials in all groups. This suggests that due to the complexity of our selected task, implicit and explicit strategies were not as easily dissociable as in typical studies of motor learning strategies, like the classic visuomotor rotation task, which are limited to fewer degrees of freedom \cite{Maresch2021MeasuresDont}\cite{Krakauer2000LearningTrajectories}.

\subsection{Performance Dependence on Biomechanics}

The difference in rotational accuracy between groups was most pronounced during sections of the path requiring clockwise wrist motion. Participants in the control group especially tended to under-rotate during clockwise motion, which corresponded to supination of the forearm. This is as opposed to the pronation motion required by counterclockwise rotation (Fig.~\ref{angle_vs_error}B). 

While the pronation and supination exist on the same axis, the recruitment of muscles and geometry of the bones of the arm are vastly different between the two \cite{Gordon2004ElectromyographicAdults}. Previous work has demonstrated that while gripping, the loading on flexor muscles of the forearm is highest in a supinated arm posture, as opposed to pronated or neutral \cite{Mogk2003TheGripping}. 

Additionally, in the context of our task, the initial movement of the ring was one that required a counterclockwise, or pronated motion, but most participants initially grasped the stylus in a neutral position. This likely led to ``counterclockwise" path sections occurring with a neutral or lightly pronated wrist angle, while ``clockwise" path sections required a more extreme supination. Irrespective of the different configuration of muscular requirements and joint loading of extremes of either forearm pronation and supination, humans naturally avoid maximal movements in either direction, spending 50\% of their time in the central 20\% of the functional range of motion in daily living \cite{Anderton2023MovementLiving}. 

\subsection{Limitations and Future Work}
We demonstrated the advantages of motor learning under a training paradigm influenced by principles of implicit motor learning, but we did not separate the relative contributions of the underlying implicit and explicit learning processes between our experimental groups. This differentiation is difficult to accomplish in a complex task, as commonly accepted methods of quantifying implicit versus explicit motor learning processes are used and validated in simple tasks like point-to-point reaching \cite{Maresch2021MeasuresDont}. This separation may be further explored in complex tasks in future work through assessment of conscious strategy through methods like asking subjects to pantomime learned paths with their opposite hand, with the assumption that the difference between self-reported knowledge and actual task performance is implicitly-learned strategy \cite{Albert2021AnAdaptation}\cite{Maresch2021MeasuresDont}\cite{McDougle2019DissociableLearning}. Additionally, to dissociate implicit and explicit strategies more successfully through multitasking trials, participants could be constrained to task completion during multitasking within the same time as their average during the end of the training period to force the abandonment of conscious strategies.

In this study, the post-training assessment was limited to participant performance directly after the conclusion of the training period. Previous work in the field of procedural learning has suggested that implicitly learned motor tasks have superior long-term retention as compared to tasks learned through primarily explicit mechanisms \cite{Krakauer2019MotorLearning}, although this retention is not necessarily permanent \cite{Willingham1997Long-termDelay}. Retesting participants after days or weeks would probe the retention of motor skills developed through the real-time feedback paradigm.

The results of this work, specifically the translational component of the task, were limited by the design of the study; specifically, the large size of the ring made the translational component straightforward. The task could be made less intuitive to learn so that participants must rely more upon the feedback provided. Additionally, because we sought to examine single-task adaptation under differential feedback paradigms, the wire path remained the same throughout the course of the experiment. This resulted in participants adapting to one path configuration out of infinite possibilities. Future work may seek to explore how differentiation in feedback timing leads to generalizable skills in RMIS performance instead of adaptation to a particular task. \responsetext{Additionally, we observed that participants in the real-time feedback group appeared to outperform those in the replay feedback and control groups in the translational component of the task in sections with more rapid changes in curvature. However, this result was preliminary and not formally tested in our study design; future work could explore and solidify this association.}

Finally, the cause for the demonstrated correlation between feedback condition and willingness to supinate the forearm is uncertain. Future work could have the initial grasp of the stylus correspond to a neutral ring position and require equal pronation and supination demands throughout the course of the task, to detect whether this finding was the result of a true preference for pronation movements or just a reluctance for any rotational extremes.
 
\section{Conclusions}
We demonstrated that a training paradigm inspired by the principles of implicit motor learning led to reduction in task error after training as compared to a paradigm inspired by explicit motor learning. Our implicitly-inspired training involved the provision of multi-sensory error feedback to subjects in real-time, while explicitly-inspired training provided that same feedback after task performance. We correctly predicted the superior performance effects of training inspired by implicit motor learning in the context of ring orientation, but found that training condition did not have an overall effect on accuracy of position. However, when examining more difficult sections of the path (i.e., areas of tight curves requiring rapid movements), participants who trained with real-time feedback had superior performance in positional accuracy. Additionally, participants receiving real-time feedback were more likely to rotate their wrists into an uncomfortable position to achieve correct orientation than those receiving replay or no feedback, indicating a greater attention to task requirements.

The results of this study have clinical relevance due to the state-of-the-art methodology in surgical training. Typically, surgeons are given post hoc feedback about performance after the completion of a task, similar to our replay feedback condition. Our study indicates that a shift to emphasis on real-time feedback during training will lead to more rapid skill acquisition. Furthermore, real-time feedback may lead to trainees having greater attention to detail and accuracy in challenging components of a task, which is critical in a surgical setting. \responsetext{Additionally, the findings involving reluctance towards wrist over-supination have implications for ergonomic design of RMIS systems; designers should note that accuracy falters at these extremes of rotation.} These results mark a step towards optimization of training for RMIS and have the potential to improve surgeon performance and patient outcomes.

\bibliographystyle{IEEEtran}
\bibliography{references_etalCTL}

\begin{thebibliography}{10}
\providecommand{\url}[1]{#1}
\csname url@samestyle\endcsname
\providecommand{\newblock}{\relax}
\providecommand{\bibinfo}[2]{#2}
\providecommand{\BIBentrySTDinterwordspacing}{\spaceskip=0pt\relax}
\providecommand{\BIBentryALTinterwordstretchfactor}{4}
\providecommand{\BIBentryALTinterwordspacing}{\spaceskip=\fontdimen2\font plus
\BIBentryALTinterwordstretchfactor\fontdimen3\font minus \fontdimen4\font\relax}
\providecommand{\BIBforeignlanguage}[2]{{%
\expandafter\ifx\csname l@#1\endcsname\relax
\typeout{** WARNING: IEEEtran.bst: No hyphenation pattern has been}%
\typeout{** loaded for the language `#1'. Using the pattern for}%
\typeout{** the default language instead.}%
\else
\language=\csname l@#1\endcsname
\fi
#2}}
\providecommand{\BIBdecl}{\relax}
\BIBdecl

\bibitem{Maeso2010EfficacyMeta-analysis}
S.~Maeso \emph{et~al.}, ``{Efficacy of the da Vinci surgical system in abdominal surgery compared with that of laparoscopy: A systematic review and meta-analysis},'' \emph{Annals of Surgery}, vol. 252, no.~2, pp. 254--262, 2010.

\bibitem{Mjaess2023NewNow}
G.~Mjaess, L.~Orecchia, and S.~Albisinni, ``{New robotic platforms for prostate surgery: The future is now},'' \emph{Prostate Cancer and Prostatic Diseases}, vol.~26, no.~3, pp. 519--520, 2023.

\bibitem{Munz2004TheModels}
Y.~Munz \emph{et~al.}, ``{The benefits of stereoscopic vision in robotic-assisted performance on bench models},'' \emph{Surgical Endoscopy and Other Interventional Techniques}, vol.~18, no.~4, pp. 611--616, 2004.

\bibitem{Huffman2021AreProficiency}
E.~M. Huffman \emph{et~al.}, ``{Are current credentialing requirements for robotic surgery adequate to ensure surgeon proficiency?}'' \emph{Surgical Endoscopy}, vol.~35, no.~5, pp. 2104--2109, 2021.

\bibitem{Nisky2014EffectsSurgeons}
I.~Nisky, A.~M. Okamura, and M.~H. Hsieh, ``{Effects of robotic manipulators on movements of novices and surgeons},'' \emph{Surgical Endoscopy}, vol.~28, no.~7, pp. 2145--2158, 2014.

\bibitem{Tausch2012ContentTracker}
T.~J. Tausch \emph{et~al.}, ``{Content and construct validation of a robotic surgery curriculum using an electromagnetic instrument tracker},'' \emph{Journal of Urology}, vol. 188, no.~3, pp. 919--923, 2012.

\bibitem{Sharon2021RateEvaluation}
Y.~Sharon \emph{et~al.}, ``{Rate of orientation change as a new metric for robot-assisted and open surgical skill evaluation},'' \emph{IEEE Transactions on Medical Robotics and Bionics}, vol.~3, no.~2, pp. 414--425, 2021.

\bibitem{Enayati2018RoboticStudy}
N.~Enayati \emph{et~al.}, ``{Robotic Assistance-as-Needed for Enhanced Visuomotor Learning in Surgical Robotics Training: An Experimental Study},'' in \emph{IEEE Int'l Conf. on Robotics and Automation}, 2018, pp. 6631--6636.

\bibitem{Oquendo2024HapticEnvironment}
Y.~A. Oquendo \emph{et~al.}, ``{Haptic guidance and haptic error amplification in a virtual surgical robotic training environment},'' \emph{IEEE Transactions on Haptics}, vol.~17, no.~3, pp. 417--428, 2024.

\bibitem{Coad2017TrainingSystem}
M.~M. Coad \emph{et~al.}, ``{Training in divergent and convergent force fields during 6-DOF teleoperation with a robot-assisted surgical system},'' in \emph{IEEE World Haptics Conference}, 2017, pp. 195--200.

\bibitem{Avraham2020TheAdaptation}
C.~Avraham and I.~Nisky, ``{The effect of tactile augmentation on manipulation and grip force control during force-field adaptation},'' \emph{Journal of NeuroEngineering and Rehabilitation}, vol.~17, no.~1, 2020.

\bibitem{Quek2019EvaluationTasks}
Z.~F. Quek, W.~R. Provancher, and A.~M. Okamura, ``{Evaluation of Skin Deformation Tactile Feedback for Teleoperated Surgical Tasks},'' \emph{IEEE Transactions on Haptics}, vol.~12, no.~2, pp. 102--113, 2019.

\bibitem{Sullivan2022HapticTask}
J.~L. Sullivan \emph{et~al.}, ``{Haptic Feedback Based on Movement Smoothness Improves Performance in a Perceptual-Motor Task},'' \emph{IEEE Transactions on Haptics}, vol.~15, no.~2, pp. 382--391, 2022.

\bibitem{Krakauer2019MotorLearning}
J.~W. Krakauer \emph{et~al.}, ``{Motor Learning},'' \emph{Comprehensive Physiology}, vol.~9, no.~2, pp. 613--663, 2019.

\bibitem{Masters2019AdvancesLearning}
R.~S. Masters, T.~van Duijn, and L.~Uiga, ``{Advances in implicit motor learning},'' in \emph{Skill Acquisition in Sport: Research, Theory and Practice}, 3rd~ed.\hskip 1em plus 0.5em minus 0.4em\relax Routledge, 2019, pp. 77--96.

\bibitem{Albert2021AnAdaptation}
S.~T. Albert \emph{et~al.}, ``{An implicit memory of errors limits human sensorimotor adaptation},'' \emph{Nature Human Behaviour}, vol.~5, no.~7, pp. 920--934, 2021.

\bibitem{Mazzoni2006AnAdaptation}
P.~Mazzoni and J.~W. Krakauer, ``{An implicit plan overrides an explicit strategy during visuomotor adaptation},'' \emph{Journal of Neuroscience}, vol.~26, no.~14, pp. 3642--3645, 2006.

\bibitem{Masters2008ImplicitMulti-tasking}
R.~S. Masters \emph{et~al.}, ``{Implicit motor learning in surgery: Implications for multi-tasking},'' \emph{Surgery}, vol. 143, no.~1, pp. 140--145, 2008.

\bibitem{El-Kishawi2021EffectEndodontics}
M.~El-Kishawi \emph{et~al.}, ``{Effect of errorless learning on the acquisition of fine motor skills in pre-clinical endodontics},'' \emph{Australian Endodontic Journal}, vol.~47, no.~1, pp. 43--53, 2021.

\bibitem{Shadmehr1998Time-dependentSubjects}
R.~Shadmehr, J.~Brandt, and S.~Corkin, ``{Time-dependent motor memory processes in amnesic subjects},'' \emph{Journal of Neurophysiology}, vol.~80, no.~3, pp. 1590--1597, 1998.

\bibitem{Telgen2014MirrorNovo}
S.~Telgen, D.~Parvin, and J.~Diedrichsen, ``{Mirror reversal and visual rotation are learned and consolidated via separate mechanisms: Recalibrating or learning de novo?}'' \emph{Journal of Neuroscience}, vol.~34, no.~41, pp. 13\,768--13\,779, 2014.

\bibitem{Nisky2014UncontrolledNovices}
I.~Nisky, M.~H. Hsieh, and A.~M. Okamura, ``{Uncontrolled manifold analysis of arm joint angle variability during robotic teleoperation and freehand movement of surgeons and novices},'' \emph{IEEE Transactions on Biomedical Engineering}, vol.~61, no.~12, pp. 2869--2881, 2014.

\bibitem{Okamura2009HapticSurgery}
A.~M. Okamura, ``{Haptic feedback in robot-assisted minimally invasive surgery},'' \emph{Current Opinion in Urology}, vol.~19, no.~1, pp. 102--107, 2009.

\bibitem{Moorthy2003ObjectiveSurgery}
K.~Moorthy \emph{et~al.}, ``{Objective assessment of technical skills in surgery},'' \emph{British Medical Journal}, vol. 327, no. 7422, pp. 1032--1037, 2003.

\bibitem{Rahimi2024TrainingMethods}
A.~M. Rahimi \emph{et~al.}, ``{Training in robotic-assisted surgery: a systematic review of training modalities and objective and subjective assessment methods},'' \emph{Surgical Endoscopy}, vol.~38, no.~7, pp. 3547--3555, 2024.

\bibitem{Smith2014FundamentalsDevelopment}
R.~Smith, V.~Patel, and R.~Satava, ``{Fundamentals of robotic surgery: A course of basic robotic surgery skills based upon a 14-society consensus template of outcomes measures and curriculum development},'' \emph{International Journal of Medical Robotics and Computer Assisted Surgery}, vol.~10, no.~3, pp. 379--384, 2014.

\bibitem{Schween2014OnlineRotation}
R.~Schween \emph{et~al.}, ``{Online and post-trial feedback differentially affect implicit adaptation to a visuomotor rotation},'' \emph{Experimental Brain Research}, vol. 232, no.~9, pp. 3007--3013, 2014.

\bibitem{Schween2017FeedbackRotation}
R.~Schween and M.~Hegele, ``{Feedback delay attenuates implicit but facilitates explicit adjustments to a visuomotor rotation},'' \emph{Neurobiology of Learning and Memory}, vol. 140, pp. 124--133, 2017.

\bibitem{Poolton2016MultitaskLearning}
J.~M. Poolton \emph{et~al.}, ``{Multitask training promotes automaticity of a fundamental laparoscopic skill without compromising the rate of skill learning},'' \emph{Surgical Endoscopy}, vol.~30, no.~9, pp. 4011--4018, 2016.

\bibitem{Kleynen2014UsingLearning}
M.~Kleynen \emph{et~al.}, ``{Using a Delphi technique to seek consensus regarding definitions, descriptions and classification of terms related to implicit and explicit forms of motor learning},'' \emph{PLoS ONE}, vol.~9, no.~6, e100227, 2014.

\bibitem{McDougle2015ExplicitLearning}
S.~D. McDougle, K.~M. Bond, and J.~A. Taylor, ``{Explicit and implicit processes constitute the fast and slow processes of sensorimotor learning},'' \emph{Journal of Neuroscience}, vol.~35, no.~26, pp. 9568--9579, 2015.

\bibitem{Maresch2021MeasuresDont}
J.~Maresch, L.~Mudrik, and O.~Donchin, ``{Measures of explicit and implicit in motor learning: What we know and what we don't},'' \emph{Neuroscience and Biobehavioral Reviews}, vol. 128, pp. 558--568, 2021.

\bibitem{Krakauer2000LearningTrajectories}
J.~W. Krakauer \emph{et~al.}, ``{Learning of visuomotor transformations for vectorial planning of reaching trajectories},'' \emph{Journal of Neuroscience}, vol.~20, no.~23, pp. 8916--8924, 2000.

\bibitem{Gordon2004ElectromyographicAdults}
K.~D. Gordon \emph{et~al.}, ``{Electromyographic activity and strength during maximum isometric pronation and supination efforts in healthy adults},'' \emph{Journal of Orthopaedic Research}, vol.~22, no.~1, pp. 208--213, 2004.

\bibitem{Mogk2003TheGripping}
J.~P. Mogk and P.~J. Keir, ``{The effects of posture on forearm muscle loading during gripping},'' \emph{Ergonomics}, vol.~46, no.~9, pp. 956--975, 2003.

\bibitem{Anderton2023MovementLiving}
W.~Anderton \emph{et~al.}, ``{Movement preferences of the wrist and forearm during activities of daily living},'' \emph{Journal of Hand Therapy}, vol.~36, no.~3, pp. 580--592, 2023.

\bibitem{McDougle2019DissociableLearning}
S.~D. McDougle and J.~A. Taylor, ``{Dissociable cognitive strategies for sensorimotor learning},'' \emph{Nature Communications}, vol.~10, 40, 2019.

\bibitem{Willingham1997Long-termDelay}
D.~B. Willingham and J.~A. Dumas, ``{Long-term retention of a motor skill: Implicit sequence knowledge is not retained after a one-year delay},'' \emph{Psychological Research}, vol.~60, no. 1-2, pp. 113--119, 1997.

\end{thebibliography}

\end{document}